\documentclass[runningheads]{llncs}

\usepackage{eccv}

\usepackage{marvosym}

\makeatletter
\def\customsymbol#1{
    \ifcase\number\value{#1}
        \or *
        \or \Letter
    \else\@ctrerr
    \fi
}
\makeatother

\usepackage{eccvabbrv}

\usepackage{graphicx}
\usepackage{booktabs}
\usepackage{multirow}
\usepackage{subcaption}
\usepackage{color}
\usepackage[linesnumbered,ruled,vlined]{algorithm2e}
\usepackage[accsupp]{axessibility}  % Improves PDF readability for those with disabilities.

\newcommand{\LH}[1]{#1}
\newcommand{\remove}[1]{\iffalse
\fi}
\usepackage[pagebackref,breaklinks,colorlinks,citecolor=eccvblue]{hyperref}
\usepackage{orcidlink}
\newcommand{\grad}[2]{\frac{\partial{#1}}{\partial{#2}}}

\begin{document}

\title{Fisheye-GS: Lightweight and Extensible Gaussian Splatting Module for Fisheye Cameras} 


% Second Author\inst{2,3}\orcidlink{1111-2222-3333-4444} \and
% Third Author\inst{3}\orcidlink{2222--3333-4444-5555}}
\author{Zimu Liao\inst{1} \and Siyan Chen\inst{1} \and Rong Fu\inst{1,\dagger} \and Yi Wang\inst{1} \and Zhongling Su\inst{1,\ddagger} \and Hao Luo\inst{1} \and Li Ma\inst{1\and3} \and Linning Xu\inst{2} \and Bo Dai\inst{1} \and Hengjie Li\inst{1} \and Zhilin Pei\inst{1} \and Xingcheng Zhang\inst{1}}

\authorrunning{Z.~Liao et al.}
\titlerunning{Fisheye-GS: GS Module for Fisheye Cameras}

\institute{Shanghai Artificial Intelligence Laboratory  \and The Chinese University of Hong Kong \and Hong Kong University of Science and Technology \\
$^{\dagger}$furong@pjlab.org.cn \quad
$^{\ddagger}$suzhongling@pjlab.org.cn}

\maketitle

\setcounter{footnote}{0}
\renewcommand{\thefootnote}{\fnsymbol{footnote}}

\begin{abstract}
  Recently, 3D Gaussian Splatting (3DGS) has garnered attention for its high fidelity and real-time rendering. However, adapting 3DGS to different camera models, particularly fisheye lenses, poses challenges due to the unique 3D to 2D projection calculation. Additionally, there are inefficiencies in the tile-based splatting, especially for the extreme curvature and wide field of view of fisheye lenses, 
which are crucial for its broader real-life applications. To tackle these challenges, we introduce Fisheye-GS\footnote{Available at \href{https://github.com/zmliao/Fisheye-GS}{https://github.com/zmliao/Fisheye-GS}}. This innovative method recalculates the projection transformation and its gradients for fisheye cameras. Our approach can be seamlessly integrated as a module into other efficient 3D rendering methods, emphasizing its extensibility, lightweight nature, and modular design. Since we only modified the projection component, it can also be easily adapted for use with different camera models. Compared to methods that train after undistortion, our approach demonstrates a clear improvement in visual quality.
\remove{Our method reduces computational redundancy and memory usage, significantly improving rendering speed and enabling 3DGS to be effectively used with a wider field of view.}\remove{With our optimization, we achieved a maximum rendering efficiency of 570 FPS at a high-definition resolution of 1752 $\times$ 1168.}
\end{abstract}

\section{Introduction}
\label{sec:intro}
%\SH{PJLAB}
The advent of high-resolution fisheye cameras has transformed how we capture and interact with the world, providing a 180° or greater field of view that offers unique perspectives. These perspectives are essential for various real-world applications. In immersive virtual reality (VR), fisheye cameras enable the creation of panoramic environments that enhance the user experience by providing a more comprehensive and realistic field of vision \cite{xiong1997creating}. This is crucial for applications such as VR gaming \cite{Bourke2009idome}, virtual tourism \cite{zhang2016meusem}, and training simulations \cite{Jakab2024simulation}, where a wide and uninterrupted view can significantly improve immersion and usability. In surveillance systems, fisheye cameras allow for broader area coverage with fewer cameras, reducing costs and complexity. They are particularly valuable in security and monitoring scenarios, such as in large public spaces\cite{konrad2024high}, retail stores \cite{vandewiele2012visibility, yang2023epformer}, and traffic monitoring \cite{deng2017cnn}, where capturing every angle with great detail is vital for effective oversight and incident management.
\begin{figure}[htb]
    \centering
    \includegraphics[width=\linewidth]{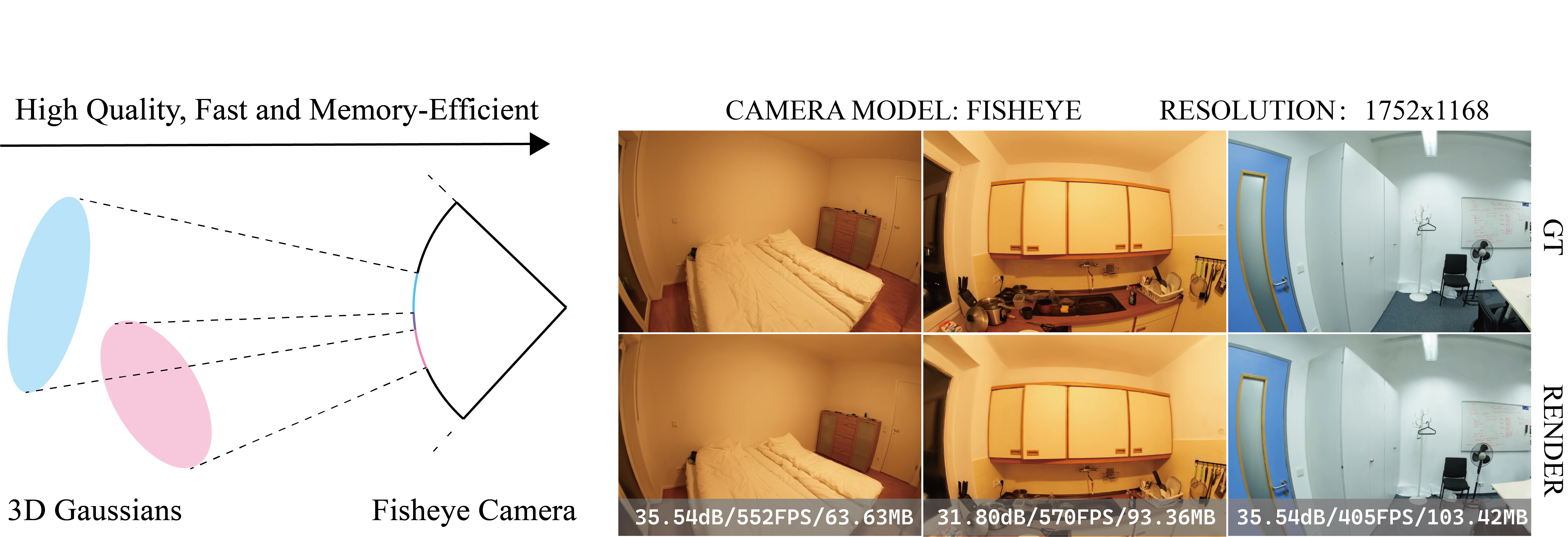}
    \caption{Our Fisheye-GS. We have directly trained our 3DGS\cite{3DGS} model from images captured from fisheye cameras without undistortion to pinhole cameras. We then integrate our Fisheye-GS as a lightweight module within FlashGS \cite{feng2024flashgsefficient3dgaussian}, an efficient rendering technique for 3DGS, to evaluate its visual quality and performance. }
    \label{teaser}
\end{figure}
However, integrating fisheye imagery into novel view synthesis poses several challenges due to the inherent curvature and distortion. While Neural Radiance Fields (NeRF)~\cite{Mildenhall2020NeRFRS} handles the distortions of fisheye lenses through its ray sampling approach, which can easily accommodate various camera models, the state-of-the-art 3D Gaussian Splatting (3DGS)~\cite{3DGS} faces difficulties. 
The original 3DGS approach relies on perspective projections. This is less adept at managing the extreme curvature and wide field of view of fisheye imagery, leading to a lack of accuracy and efficiency in rendering.

In this work, we aim to modularize the lightweight integration of fisheye cameras with 3D Gaussian Splatting (3DGS) to fully leverage the potential of fisheye lenses in real-world applications requiring real-time rendering performance. By combining the wide field of view of fisheye lenses with the efficient rendering capabilities of 3DGS, we seek to address the challenges posed by the curvature and distortion inherent in fisheye imagery.

Our proposed method, Fisheye-GS, is the first work that seamlessly integrates fisheye camera data into 3D Gaussian Splatting, unlocking new possibilities for applications in virtual reality, gaming, security, and more. We achieve this by:
\begin{itemize}
\item We first directly calculate the equidistant projection transformation and derive the means and covariance of Gaussians for fisheye lenses. It is crucial for maintaining image quality and precision across the extreme curvature and wide field of view.
\item We then formulate the gradients required for optimization in the training phase and develop corresponding CUDA implementations, ensuring that the 3DGS model learns efficiently and produces distortion-free imagery.

\item Finally, we only modified the projection model. This lightweight approach can be effectively modularized for integration into high-performance 3DGS methods and can be extended to accommodate other camera models.
\remove{Finally, we delve into eliminating computational redundancies and optimizing the CUDA kernel to greatly improve GPU performance which ensures that the proposed method is well-suited for real-time applications.}
\end{itemize}

Extensive experiments and comparisons with existing methods demonstrate the effectiveness of Fisheye-GS in producing high-quality, distortion-free imagery, making it ideal for various 3D graphics applications. 
%In our knowledge, 
Up to our best knowledge, Fisheye-GS is the first open-source project that supports both training and rendering of 3DGS on fisheye cameras with efficient speed and performance. %up to now.

\section{Related Work} % Remove NeRF

\subsection{Novel View Synthesis} % 3D Gaussian Splatting
\paragraph{Neural Radiance Fields (NeRF).}
NeRF\cite{Mildenhall2020NeRFRS} is a groundbreaking technique in novel view synthesis, enabling the generation of realistic images of 3D scenes from any viewpoint. NeRF uses a deep neural network to encode a continuous volumetric scene function, which maps 3D points and viewing directions to corresponding RGB color and volume density values.
To improve rendering quality, some methods refine the point sampling strategy during ray marching, leading to more accurate modeling of the volume rendering process \cite{Xu2022PointNeRFPN}. Others enhance rendering by reparameterizing the scene representation, creating a more compact encoding, and simplifying the learning process \cite{Barron2021MipNeRF3U, Barron2021MipNeRFAM}.
Despite these optimizations, NeRF remains computationally intensive during rendering. Typically, NeRF is represented as a global MLP encoding the entire scene space, which can be inefficient and costly for complex or large-scale scenes. To address this, some works have explored alternative scene representations, such as voxel grids and octree-based approaches \cite{Yu2021PlenoxelsRF, Yu2021PlenOctreesFR}. While these methods improve rendering speed, achieving real-time performance is still challenging due to the inherent ray marching strategy in volume rendering.
% NeRF\cite{Mildenhall2020NeRFRS} represents a breakthrough in novel view synthesis, which aims to generate realistic images of a 3D scene from any viewpoint. NeRF employs a fully connected deep neural network to encode a continuous volumetric scene function. This function maps a 3D point and a viewing direction to the corresponding RGB color and volume density. To enhance rendering quality, some methods directly refine the point sampling strategy in ray marching, resulting in more accurate modeling of the volume rendering process \cite{Xu2022PointNeRFPN}. Others enhance rendering by reparameterizing the scene, which creates a more compact scene representation and simplifies the learning process \cite{Barron2021MipNeRF3U, Barron2021MipNeRFAM}. Despite many optimizations, NeRF still consumes much time and computational power during the rendering process. NeRFs are typically represented as global MLPs encoding the entire scene space. However, this approach can be inefficient and costly when reconstructing complex and large-scale scenes. Some works focus on innovations in scene representations.  They propose voxel grids \cite{Yu2021PlenoxelsRF} and octree-based \cite{Yu2021PlenOctreesFR}. Although these methods greatly enhance rendering speed, achieving real-time rendering remains difficult due to the inherent ray marching strategy in volume rendering.
% \TODO{Add NeRF in ONE sentence}
\paragraph{3D Gaussian Splatting}
3D Gaussian Splatting (3DGS)~\cite{3DGS} models the scene with a set of Gaussians, allowing efficient rendering through rasterizing these Gaussians into images. While 3DGS produces high-quality reconstruction results, there is room for improvement\remove{ in its rendering capabilities}. \cite{Zhu2023FSGSRF,Xiong2023SparseGSR3,Li2024DNGaussianOS,Charatan2023pixelSplat3G,Chen2024MVSplatE3} are designed to improve rendering performance under challenging inputs like sparse views. These methods enhance the robustness and accuracy of the rendered images in difficult scenarios. For anti-aliasing\remove{ and improving both efficiency and quality}, approaches like \cite{Yu2023MipSplattingA3,Yan2023MultiScale3G,Liang2024AnalyticSplattingA3} are utilized.\remove{ These techniques help reduce aliasing artifacts and ensure smooth transitions in the rendered images, providing a more visually appealing output.} Anchor-based methods, such as \cite{Lu2023ScaffoldGSS3,Ren2024OctreeGSTC} focus on improving efficiency and quality using hierarchical structures and anchor points. \cite{Ye_gsplat, feng2024flashgsefficient3dgaussian} has proposed a suite of optimization strategies to amplify the performance of the rasterizing process in 3DGS. These methods enhance the computational efficiency and scalability of the rendering process\remove{, allowing for faster and more accurate scene reconstructions}.

\subsection{Various Camera Models for Radiance Fields} 
\LH{Since NeRF cast rays from the camera center and samples points in 3D space for volume rendering, it is flexible in handling various camera models, as also integrated in ~\cite{TanciWNLYKWKASAMK2023}, enables the reconstruction of a wide range of scenes from different types of input images. A few striking examples of models can be SC-NeRF~\cite{Jeong2021SelfCalibratingNR} for reconstruct 3D scenes and camera parameters without traditional calibration objects, 360FusionNeRF~\cite{Kulkarni2022360FusionNeRFPN} for 360-degree panorama rendering, OmniNeRF~\cite{Shen2022OmniNeRFHO} for enhancing 3D reconstruction at scene edges, PERF~\cite{Wang2023PERFPN} for demonstrating the superiority of extensive 3D roaming, and Neuro Lens~\cite{Xian2023NeuralLM} for end-to-end optimization of the image formation process.}

As for 3DGS, it primarily works in pinhole cameras. Adapting it to cameras with distortions is challenging because the projection of covariance is more difficult to compute. Some approaches have tried to extend 3DGS for specific cameras with distortion. \remove{with fisheye and panoramic camera models involves additional considerations to handle specific distortions and projections associated with these images involving various approximations in the rasterization process.}\remove{\cite{yeshwanthliu2023scannetpp} suggests pre-processing fisheye images to correct for radial distortion, effectively converting them to a more standard perspective view before applying Gaussian Splatting. Alternatively, one can modify the projection process within the Gaussian Splatting pipeline to approximate the non-linear mapping, ensuring correct positioning and scaling of splats to fix the distortion.}For example, \cite{Bai2024360GSLP} projects 3D Gaussians onto the tangent plane of the unit sphere, addressing challenges in spherical projections and enhancing rendering quality by leveraging structural priors from panoramic layouts. \cite{huang2024erroranalysis3dgaussian} introduces an optimal projection strategy for various cameras by analyzing and minimizing projection errors by projecting Gaussians into a unit sphere.

\section{Preliminaries}

\subsection{3D Gaussian Splatting} %Simplify
\paragraph{Definition.}
3DGS represents a 3D model as a set of anisotropic 3D Gaussian functions and renders images via a differentiable tile-based rasterizer. Each Gaussian function is designated as the mean $\boldsymbol{\mu}\in\mathbb{R}^3$, anisotropic covariance $\mathbf{\Sigma}\in\mathbb{R}^{3\times3}$, % opacity $\sigma\in\mathbb{R}$ and view-dependent color $\mathbf{c}\in\mathbb{R}^3$ parameterized by spherical harmonics:
\begin{equation}
    G(\mathbf{x}) = e^{-\frac{1}{2}(\mathbf{x}-\boldsymbol{\mu})^T\mathbf{\Sigma}^{-1}(\mathbf{x}-\boldsymbol{\mu})},
    \label{3D Gaussian}
\end{equation}
where $\mathbf{x}$ is an arbitrary position in the world coordinate system. To constrain the positive semi-definite form, the covariance is defined as $\mathbf{\Sigma} = \mathbf{R}\mathbf{S}\mathbf{S}^T\mathbf{R}^T$, where $\mathbf{R}\in SO(3)$ is the rotation matrix represented by quaternions and $\mathbf{S}=\text{diag}(s_1,s_2,s_3)$ is the scaling matrix.

\paragraph{Transformation.} 
To project the Gaussian into the 2D plane, 3DGS applies the viewing transformation from the camera coordinate system
$\varphi\in\mathbb{R}^3\to\mathbb{R}^3$ and the projective transformation $\phi\in\mathbb{R}^3\to\mathbb{R}^2$. We can model $\varphi$ as the linear transformation $\varphi(\mathbf{x})=\mathbf{W}\mathbf{x}+\mathbf{b}$ where $\mathbf{W}\in\mathbb{R}^{3\times3}$ and $\mathbf{b}\in\mathbb{R}^3$. Then the projected mean $\boldsymbol{\mu}_p\in\mathbb{R}^2$ and covariance $\mathbf{\Sigma_p}\in\mathbb{R}^{2\times2}$ are given as follows:
\begin{equation}
    \boldsymbol{\mu}_p = \phi(\boldsymbol{\mu}_c) = \phi(\mathbf{W}\boldsymbol{\mu} + \mathbf{b}),\quad\mathbf{\Sigma_p} = \mathbf{J}\mathbf{W}\mathbf{\Sigma}\mathbf{W}^T\mathbf{J}^T,
\end{equation}
where $\boldsymbol{\mu}_c\in\mathbb{R}^3$ is the mean in the camera coordinate system and $\mathbf{J}\in\mathbb{R}^{2\times3}$ is the Jacobian of $\phi$. Notice that the $\mathbf{\Sigma_p}$ is the affine approximation. Then the projected 2D Gaussian can be modeled as an ellipse on the image plane.
\paragraph{Rendering.}To render the color $C\in{\mathbb{R}^3}$ of the pixel, 3DGS uses differentiable alpha blending from ellipses of $N$ Gaussians intersecting the pixel according to the depth order:
\begin{equation}
    \mathbf{C} = \sum_{i=1}^n \mathbf{c}_i\alpha_i\prod_{j=1}^{i-1}(1-\alpha_j),
\end{equation}
where $\alpha$ is evaluated by projected 2D Gaussians.
To determine intersections between pixels and 2D Gaussians, 3DGS applies the technique of dividing the image plane into several $16\times 16$ tiles. Firstly each ellipse determines its axis-aligned bounding box. Then 3DGS traverses each bounding box to store the intersections and sorts them by tile IDs and depths. Finally, each pixel in a certain tile performs alpha blending on intersected Gaussians to render its color.
% \begin{algorithm}[hbt!]
% \caption{PreprocessCUDA}\label{alg:preprocess}
% \KwIn{$P$\textcolor{blue}{(Number of Gaussians)} ,
%      $orig\_points$,$scales$,$scale\_modifier$,
%    $rotations$,$opacities$, $shs$\textcolor{blue}{(Spherical Harmonics Coefficients)},
%     $clamped$,$viewmatrix$, $projmatrix$,
%     $cam\_pos$}
% \KwOut{$radii$\textcolor{blue}{(radius)}, $points\_xy\_image$\textcolor{blue}{(Projected Points in Image Space)},
%     $depths$, $cov3Ds$\textcolor{blue}{(Covariances)}, $rgb$\textcolor{blue}{(Colors)},
%     $conic\_opacity$,$tiles\_touched$}
% \If{$idx \geq P$}{
%     \Return \;
% }

% $p\_view \gets {in\_frustum}()$ \;
% \If{$!p\_view$}{
%             \Return \;
%         }
% $p\_proj \gets {{project}}()$ \;
% $cov3D \gets {computeCov3D}()$ \;
% $cov \gets {computeCov2D}()$ \;
% $det \gets (cov.x \times cov.z - cov.y \times cov.y)$ \;
% \If{$det == 0.0$}{
%     \Return \;
% }
% $\lambda_1, \lambda_2 \leftarrow {eigenvalues}(\Sigma')$ \;
% $my\_radius \gets \lceil 3 \times \sqrt{\max(\lambda_1, \lambda_2)} \rceil$ \;
% $rect\_min,rect\_max \gets {getRect}()$ \;
% \If{$(rect\_max.x - rect\_min.x) \times (rect\_max.y - rect\_min.y) == 0$}{
%             \Return \;
%         }
% $rgb[idx] \gets {computeColorFromSH}()$ \;
% $depths[idx] \gets p\_view.z$ \;
% $radii[idx] \gets my\_radius$ \;
% $points\_xy\_image[idx] \gets point\_image$ \;
% $conic\_opacity[idx] \gets (conic, opacities[idx])$ \;
% $tiles\_touched[idx] \gets (rect\_max.y - rect\_min.y) \times (rect\_max.x - rect\_min.x)$ \;
% \end{algorithm}

\subsection{Equidistant Projection Model for Fisheye Cameras}
\label{equidistant_projection}
Fisheye cameras use wide-angle lenses with a large field of view, creating images with significant radial distortion to capture extensive scenes. \remove{These lenses project images with optical distortion at the edges, making fisheye cameras ideal for applications in target tracking and security.}
Unlike pinhole cameras, which form images on a plane, fisheye cameras form images on a spherical surface. Their projection models include equidistant, equisolid angle, orthogonal, and stereographic projections.

We denote $\theta$ as the angle of incidence, and $r_d$ as the distance from the optical center to the projection point. In equidistant projection, this distance equals the arc length on the projection plane, given by:
\begin{equation}\label{equal_distance}
r_{d} = f\theta,
\end{equation}
where $f$ is the camera's focal length. This model ensures equal distances for equal angles of incidence.

For the point $\mathbf{p\in\mathbb{R}^3}=(x_c,y_c,z_c)^T$ in camera space, the distance to the z-axis is:

\begin{equation}
l_z = \sqrt{x_c^2 + y_c^2}.
\end{equation}
The angle of incidence is %$\theta$
%\begin{equation}
$\theta = \arctan\left(\dfrac{\sqrt{x_c^2 + y_c^2}}{z_c}\right).$
%\end{equation}
Using equidistant projection, the pixel coordinates are:
\begin{equation}
x_p = c_x + \dfrac{f_x \theta x_c}{l_z}, \quad
y_p = c_y + \dfrac{f_y \theta y_c}{l_z},
\end{equation}
where $f_x$ and $f_y$ are the focal lengths in the $x$ and $y$ directions, and $(c_x, c_y)$ is the optical center's pixel coordinate.

\section{Methods} %Do not Simplify
\begin{figure}[htbp]
    \centering
    \begin{minipage}[t]{\textwidth}
         \centering
         \includegraphics[width=\textwidth]{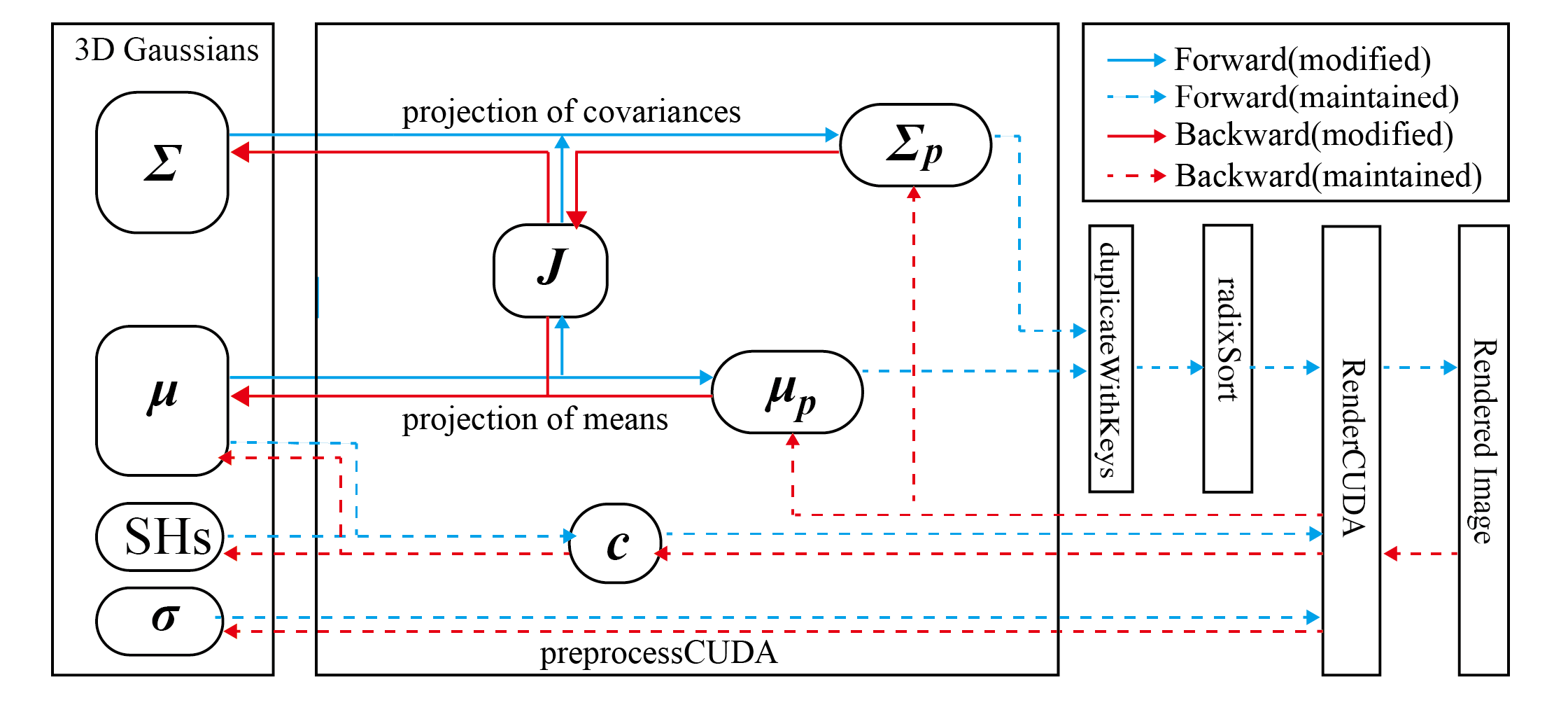}
        %\label{fig:method}
     \end{minipage}
    \caption{Analysis and Modification of 3DGS pipeline. To apply a fisheye camera for 3DGS, we derive the equidistant projection and its Jacobian matrix. Then we implement the projection as the red arrows. We have also adjusted the back-propagation illustrated in the purple arrows to align the modified projection of the fisheye cameras. Our module enables both training and rendering of 3DGS for fisheye cameras.}
    \label{fig:method}
\end{figure}
\remove{To apply fisheye cameras for 3DGS, we first derive the means and covariance of Gaussians according to equidistant projection transformation. Furthermore, the gradients for the training stage are derived in detail.}
\subsection{Analysis of the Rendering Process of 3DGS}
At first, we identify the projection-related components by analyzing the rendering process of 3D Gaussian Splatting (3DGS). As is shown in Figure \ref{fig:method}, the rendering process is mainly composed of 4 processes, geometry preprocessing, tile binning, sorting, and rasterizing. The preprocessing is implemented by \verb|preprocessCUDA|, which projects the Gaussians into a 2D surface via the projection transform $\varphi$. The tile binning implemented by \verb|duplicateWithKeys| binds the tiles overlapping the corresponding Gaussians. Then the sorting process uses \verb|radixSort| to sort Gaussians with tiles according to tile IDs and depths. Finally, \verb|renderCUDA| renders the Gaussians to image as the rasterizing process.

To apply fisheye cameras for 3DGS, we can only modify the projection transform according to equidistant projection $\phi$ in the kernel \verb|preprocessing|. This approach constrains the scope of modifications, facilitating the subsequent implementation of lightweight modules and the extension to other projection types. In detail, we can only adjust the projection of means and covariances as the red arrows and the back-propagation as purple arrows in Figure \ref{fig:method}. 

\subsection{The Modified Projection and Gradients of the Means}
\paragraph{Forward (blue arrows in Fig \ref{fig:method}).}Suppose the mean in camera space is $\boldsymbol{\mu}_c = (x_c, y_c, z_c)^T$ and the mean in pixel coordinates is $\boldsymbol{\mu}_p = (x_p, y_p)^T$, the projection of means is simply applying the equidistant projection model $\phi$ as Section \ref{equidistant_projection} shows:
\begin{equation}
    \boldsymbol{\mu}_p=\phi(\boldsymbol{\mu}_c).
\end{equation}
\paragraph{Backward (red arrows in Fig \ref{fig:method}).}Suppose the loss is $\mathcal{L}$, then the gradients of projected mean $\grad{\mathcal{L}}{\boldsymbol{\mu}'}$ can be calculated by vanilla 3DGS. To compute the mean in camera space $\grad{\mathcal{L}}{\boldsymbol{\mu}_c}$, we can apply the chain rule:
\begin{equation}
    \grad{\mathcal{\mathcal{L}}}{\boldsymbol{\mu}_c} = \grad{\boldsymbol{\mu}_p}{\boldsymbol{\mu_c}}\cdot\grad{\mathcal{L}}{\boldsymbol{\mu}_p}.
\end{equation}
Notice that $\grad{\boldsymbol{\mu}_p}{\boldsymbol{\mu_c}}$ is the Jacobian matrix of equidistant projection $\phi$, which also needs to be modified:
\begin{equation}
    \mathbf{J}_\phi = \grad{\boldsymbol{\mu}_p}{\boldsymbol{\mu_c}} = \begin{bmatrix}
        \dfrac{f_xx_c^2z_c}{l_z^2l_2} + \dfrac{y_c^2\theta}{l_z^3} & x_cy_c\left(\dfrac{f_xz_c}{l_z^2l_2}-\dfrac{\theta}{l_z^3}\right) & -\dfrac{f_xx_c}{l_2} \\
        x_cy_c\left(\dfrac{f_yz_c}{l_z^2l_2}-\dfrac{\theta}{l_z^3}\right) & \dfrac{f_yy_c^2z_c}{l_z^2l_2} + \dfrac{x_c^2\theta}{l_z^3} & -\dfrac{f_yy_c}{l_2}
    \end{bmatrix},
\end{equation}
where $l_2$ is the square length of the vector $\boldsymbol{\mu}_p$, i.e., $l_2 = x_c^2 + y_c^2 + z_c^2$. 

\subsection{The Modified Projection and Gradients of the Covariance}
\paragraph{Forward.}Using the modified Jacobian matrix above, we can compute the projected covariance via equidistant projection. Let $\mathbf{T}=\mathbf{J}\mathbf{W}\in\mathbb{R}^{2\times3}$, then we have the projected covariance:
\begin{equation}
    \mathbf{\Sigma_p} = \mathbf{T}\mathbf{\Sigma}\mathbf{T}^T.
\end{equation}
\paragraph{Backward.}As the matrix $\mathbf{T}$ is the function of $\boldsymbol{\mu}_{c}$, the projection above propagates gradients to both transformation matrix $\mathbf{T}$ and covariance $\mathbf{\Sigma}$.
GSplat \cite{ye2023mathematical} and Vanilla 3DGS implementation~\cite{3DGS} have derived the gradients above from the covariance projection by matrix calculus. Since the view transformation matrix $\mathbf{W}$ is not a function of $\boldsymbol{\mu}$, we can directly modify the partial derivative of the Jacobian matrix from Vanilla 3DGS to propagate the gradients:
\begin{equation}
    \dfrac{\partial{\mathbf{J}}}{\partial{x_c}} = \begin{bmatrix}
        \dfrac{f_xS_1}{l_2^2l_z^5} & \dfrac{f_xS_2}{l_2^2l_z^5} & \dfrac{f_xS_5}{l_2^2} \\
        \dfrac{f_yS_2}{l_2^2l_z^5} & \dfrac{f_yS_3}{l_2^2l_z^5} & \dfrac{2f_yx_cy_c}{l_2^2}
    \end{bmatrix}
\end{equation}
\begin{equation}
    \dfrac{\partial{\mathbf{J}}}{\partial{y_c}} = \begin{bmatrix}
        \dfrac{f_xS_2}{l_2^2l_z^5} & \dfrac{f_xS_3}{l_2^2l_z^5} & \dfrac{2f_yx_cy_c}{l_2^2} \\
        \dfrac{f_yS_3}{l_2^2l_z^5} & \dfrac{f_yS_4}{l_2^2l_z^5} & \dfrac{f_yS_6}{l_2^2}
    \end{bmatrix},
\end{equation}
\begin{equation}
    \dfrac{\partial{\mathbf{J}}}{\partial{z_c}} = \begin{bmatrix}
        \dfrac{f_xS_5}{l_2^2} & \dfrac{2f_xx_cy_c}{l_2^2} & \dfrac{2f_xx_cz_c}{l_2^2} \\
        \dfrac{2f_yx_cy_c}{l_2^2} & \dfrac{f_yS_6}{l_2^2} & \dfrac{2f_yy_cz_c}{l_2^2}
    \end{bmatrix},
\end{equation}
where
\begin{equation}
\begin{aligned}
    S_1 &= x_c(3S_7 + x_c^2S_8),\quad S_2 = y_c(S_7 + x_c^2S_8), \\
    S_3 &= x_c(S_7 + y_c^2S_8),\quad S_4 = y_c(3S_7 + y_c^2S_8), \\
    S_5 &= x_c^2 - y_c^2 -z_c^2,\quad S_6 = y_c^2 - x_c^2 - z_c^2 \\
    S_7 &= -l_2^2l_z^2\theta + l_2l_z^3z_c,\quad S_8 = 3l_2^2\theta - 3l_2l_zz_c - 2l_z^3z_c
\end{aligned}
\end{equation}
Finally, the mean $\boldsymbol{\mu}'$ accumulates the gradients from the backpropagation of the projection of the mean itself and its covariance to optimize the Gaussian's position. The projection of covariance also propagates gradients to the covariance itself\remove{to adjust the shape}.

\begin{figure}[h]
    \centering
    \includegraphics[width=0.9\linewidth]{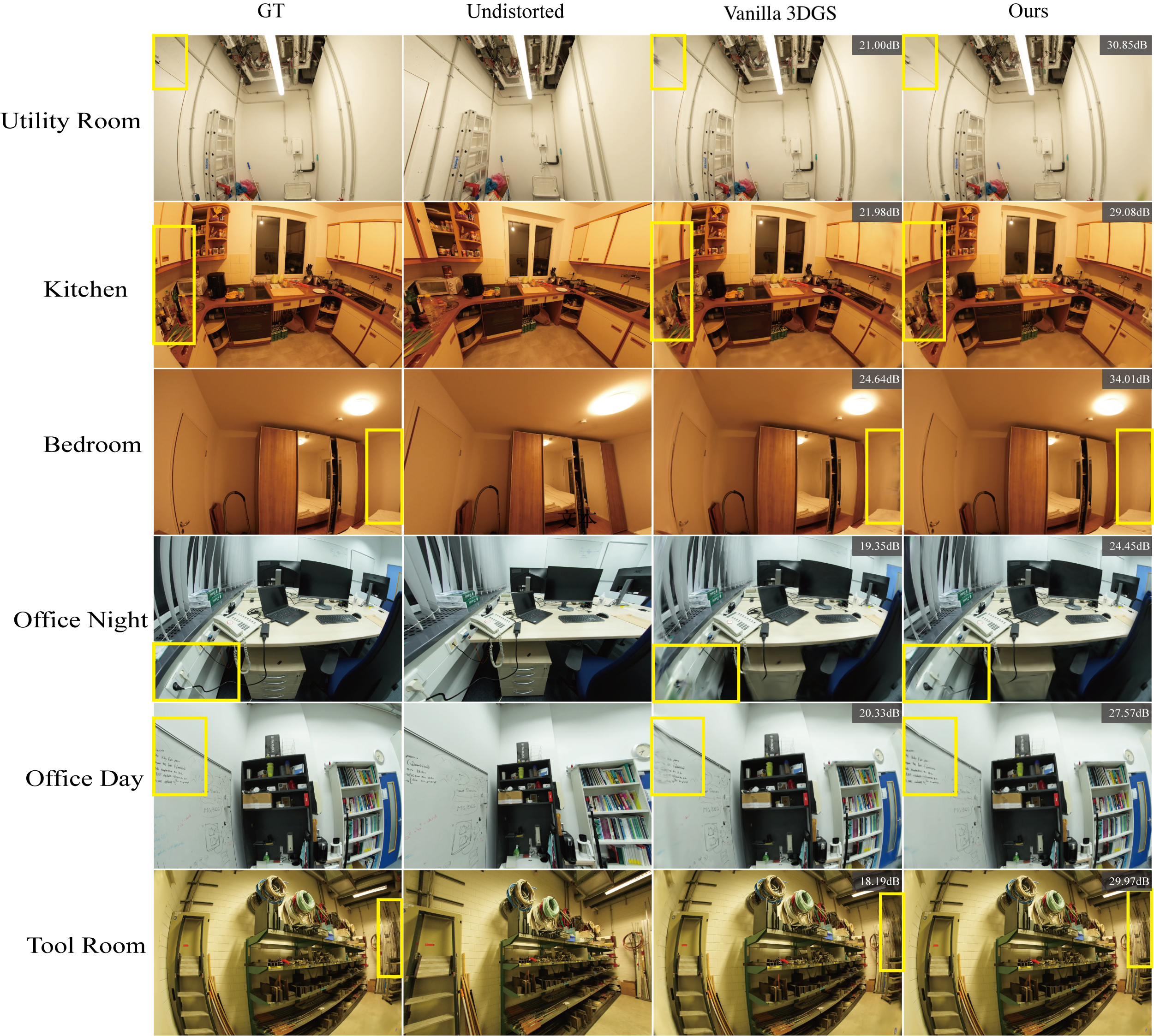}
    \caption{Qualitative comparison between Fisheye-GS and the baseline. The baseline struggles to render on edges and corners due to the clipping and interpolation from undistortion. }
    \vspace{-1em}
    \label{realfig}
\end{figure}
\begin{table}[h]
    \centering
    \begin{tabular}{c|c|cccc}
            Scene&Method&\#Gaussians&PSNR$\uparrow$&SSIM$\uparrow$&LPIPS$\downarrow$\\
         \hline
            \multirow{2}{*}{Utility Room}& Vallina 3DGS &$730320$ & $21.69$ & $0.821$ &$0.215$\\
            &Ours&$722639$&$\mathbf{26.42}$&$\mathbf{0.894}$&$\mathbf{0.177}$\\
            \hline
            \multirow{2}{*}{Big Office}&Vanilla 3DGS&$1800937$&$20.21$ & $0.72$ & $0.264$\\
            &Ours&$1442433$&$\mathbf{24.95}$&$\mathbf{0.849}$&$\mathbf{0.223}$\\
            \hline
            \multirow{2}{*}{Small Office}&Vanilla 3DGS&$712107$&$24.02$&$0.87$&$0.190$\\
            &Ours&$685902$&$\mathbf{26.15}$&$\mathbf{0.905}$&$\mathbf{0.177}$\\
            \hline
            \multirow{2}{*}{Kitchen}&Vanilla 3DGS&$704898$&$25.54$&$0.867$&$0.211$\\
            &Ours&$594699$&$\mathbf{30.63}$&$\mathbf{0.933}$&$\mathbf{0.179}$\\
            \hline
            \multirow{2}{*}{Bedroom}&Vanilla 3DGS&$355658$&$25.91$&$0.882$&$0.223$\\
            &Ours&$1442433$&$\mathbf{31.89}$&$\mathbf{0.946}$&$\mathbf{0.191}$\\
            \hline
            \multirow{2}{*}{Tool Room}&Vanilla 3DGS&$3558209$&$19.19$&$0.63$&$0.316$\\
            &Ours&$2605005$&$\mathbf{27.12}$&$\mathbf{0.855}$&$\mathbf{0.223}$\\
    \end{tabular}
    \caption{Quantitative Comparison on Scannet++ dataset. Our Fisheye-GS achieves better quality than the baseline in all scenes and is evaluated by all metrics.}
    \vspace{-1em}
    \label{realtable}
\end{table}
\section{Experiments} % Remove performance related
\subsection{Experimental Setup}
\paragraph{Dataset.} We have adopted $6$ indoor scenes from Scannet++\cite{yeshwanthliu2023scannetpp} dataset as our real-world dataset, where the images are captured by fisheye cameras and resized to the resolution of $1752\times 1168$. The dataset also provides point cloud, camera intrinsic, and poses of each view as COLMAP\cite{schonberger2016structure} SfM format. We have also generated a synthetic dataset from \cite{Mildenhall2020NeRFRS} with Blender 3.6 by equidistant fisheye camera models. We generate 200 frames with a resolution of $ 800\times800$ by rotating around the model for the Lego scene and 100 frames for other scenes. For each scene, we render the dataset separately for the field of view (FOV) settings of $120^\circ$ and $180^\circ$. For every 8 frames of the scene, the first frame is used as the test set and the others are used for training.
%We use $7/8$ of each scene as the training set\remove{ for optimization} and the rest as the test set\remove{ for evaluation}.
\paragraph{Optimization.}We have initialized Gaussians from point cloud on Scannet++ dataset as COLMAP format.\remove{ and randomly initialized $100000$ points for synthetic dataset.} For each, we train our models for $30000$ iterations and use the same hyperparameters, training strategy, and loss function as \cite{3DGS}. For the Scannet++ dataset, we randomize the background color for each iteration to model the walls of indoor scenes. Black backgrounds are used for training on synthetic datasets.
\paragraph{Baseline and Evaluation.}We compare our method against original 3DGS\cite{3DGS} as our baseline. Because 3DGS is applied for the pinhole camera model, we have undistorted the images from the Scannet++ dataset for training. The baseline shares the same training technique as our method. For a fair comparison, we evaluate the baseline from ground-truth images before undistortion, using our Fisheye-GS on both training results.
\remove{We have tested the rendering quality of our Fisheye-GS.} The evaluation resolution is the same as the training dataset. \remove{Apart from common metrics including PSNR, SSIM\cite{wang2004image} and LPIPS
\cite{zhang2018unreasonable}, we have also tested the average FPS, maximum rendering time, and memory cost of each test view on the Scannet++ dataset. The maximum number of overlaps between Gaussians and tiles are recorded to help monitor the rendering performance on the Scannet++ dataset.}For the synthetic dataset, we have tested the rendering quality in different FOVs.
% \remove{In addition, we have implemented Fisheye-GS into 2 versions called Vanilla Fisheye-GS and Efficient Fisheye-GS, the former of which does not perform redundancy elimination. We have tested both versions on the Scannet++ dataset via 3 types of GPUS with pinhole and fisheye camera models.} 

\subsection{Visual Quality Evaluation}

%Table1 : Real 
%Table2 : Performance ready
%Table3 : Synthetic ready
%Fig1 : Real 4x6
%Fig1.5 : Reason ranges? (Teaser)
%Fig2 : Synthetic 4x4 Lego Mic Hotdog Ficus GT RENDER FOV=120,180 ready
\paragraph{Evaluation on Real-world Dataset.} 
    We use visual PSNR, SSIM\cite{wang2004image} and LPIPS\cite{zhang2018unreasonable} to test our method and the baseline on 6 real indoor scenes captured by fisheye cameras. As is shown in table \ref{realtable}, our method outperforms the baseline on each metric and scene on average. Fig.~\ref{realfig} shows that the baseline performs well on the central area of the image. However, artifacts frequently appear on the edge.
    Besides the interpolation of the undistortion process leading to the loss of information, the difference between the pinhole and fisheye camera models plays a key role. The fisheye camera has a larger field of view (FOV) compared to the pinhole camera. This wide FOV leads to an undistortion process that clips the scenes at the edges of the image. Then the 3DGS struggles to learn the edge area of the scene. In other words, the corners of the scene can be more frequently viewed by our Fisheye-GS, add more geometry constraints, and show fewer artifacts in novel views.

\begin{figure}[h]
    \centering
\includegraphics[width=\linewidth]{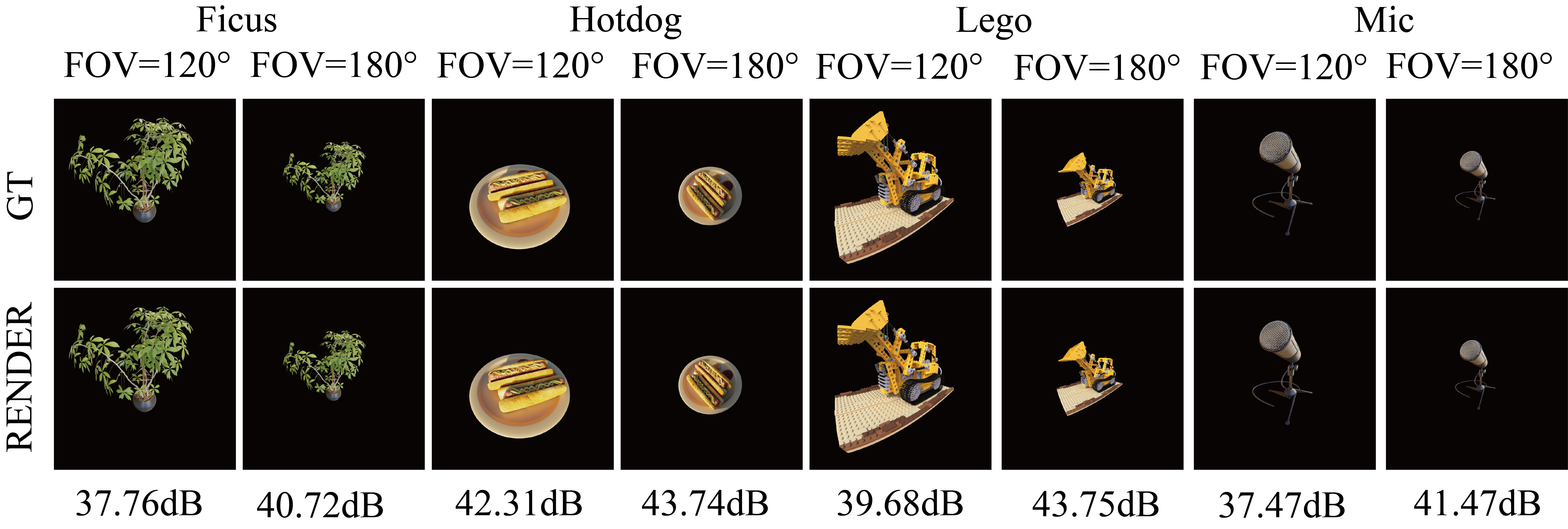}
    \caption{Result on the synthetic dataset, varying from scenes and FOVs.}
    \vspace{-1em}
    \label{syntheticf}
\end{figure}

\paragraph{Evaluation on Synthetic Dataset.}
As is shown in table \ref{synthetict} and fig \ref{syntheticf}, our Fisheye-GS can successfully model the scene from random initialization, and adapt different views. The results show our method can adapt to larger distortions like Lego in fig \ref{syntheticf}.

\begin{small}
\begin{table}[h]
    \centering
    \begin{tabular}{c|cccc|cccc}
         \multirow{2}{*}{Scene} & \multicolumn{4}{c|}{FOV=$120^{\circ}$} & \multicolumn{4}{c}{FOV=$180^{\circ}$} \\
         % \cline{2-9}
            &\#Gaussians&PSNR&SSIM&LPIPS&\#Gaussians&PSNR&SSIM&LPIPS\\
         \hline
         Lego&$144794$&$33.06$&$0.981$&$0.026$&$50357$&$30.62$&$0.9900$&$0.0141$\\
         Mic&$47773$&$43.16$&$0.997$&$0.002$&$20371$&$47.77$&$0.9986$&$0.0013$\\
         % Ship&$70939$&$37.68$&$0.973$&$0.033$&$43871$&$28.20$&$0.9426$&$0.0449$\\
        % Drum&$77208$&$33.61$&$0.988$&$0.011$&$37519$&PSNR&SSIM&LPIPS\\
         Hotdog&$62145$&$41.50$&$0.991$&$0.0128$&$37519$&$43.47$&$0.9956$&$0.0055$\\
         Ficus&$201956$&$36.90$&$0.992$&$0.0084$&$85475$&$40.01$&$0.9962$&$0.0041$\\
    \end{tabular}
    \caption{Evaluation of the synthetic dataset from different scenes and FOVs.}
    \vspace{-2em}
    \label{synthetict}
\end{table}
\end{small}

% \paragraph{Performance Evaluation on Real World Dataset.} 
% Table \ref{performance} shows the performance before and after our optimizations. With these optimizations, we achieved performance of up to 570 FPS and about $3.4\times$ speed-up ratio using an A100 GPU. As for consumer GPU RTX 4090, due to its stronger computing power but lower memory bandwidth, the acceleration effect is limited by the sort procedure. However, we still achieved good performance, meeting real-time rendering requirements.

\subsection{Discussion and Limitation}
\begin{table}[h]
    \centering
    \begin{tabular}{c|c|c|c|cccc}
         Scene & GPU & Camera & Kernel & FPS$\uparrow$ & \begin{tabular}{@{}c@{}}MaxTime\\(ms)$\downarrow$\end{tabular} & Intersects & \begin{tabular}{@{}c@{}}Memory\\(MB)$\downarrow$\end{tabular}\\
         \hline
         \multirow{8}{*}{\begin{tabular}{@{}c@{}}Small\\Office\end{tabular} } &
         \multirow{4}{*}{A100} &
         \multirow{2}{*}{Fisheye}&
         FlashGS&$405$&$4.761$&$3034430$&$103.42$\\
           &   & \ &
           Vanilla&$121$&$17.155$&$13948372$&$1062.69$\\
         \cline{3-8} &   & 
         \multirow{2}{*}{Pinhole}&
         FlashGS&$510$&$4.399$&$2415742$&$89.26$\\
           &   & \ &
           Vanilla&$125$&$17.711$&$13145841$&$1035.06$\\
         \cline{2-8} &
         % \multirow{4}{*}{V100} &
         % \multirow{2}{*}{Fisheye}&
         % Efficient&$346$&$5.121$&$3034430$&$103.42$\\
         %   &   & \ &
         %   Vanilla&$81$&$28.372$&$13948372$&$1062.69$\\
         % \cline{3-8} &   & 
         % \multirow{2}{*}{Pinhole}&
         % Efficient&$417$&$4.399$&$2415742$&$89.26$\\
         %   &   & \ &
         %   Vanilla&$125$&$17.711$&$13145841$&$1035.06$\\
         % \cline{2-8} &
         \multirow{4}{*}{RTX 4090} &
           \multirow{2}{*}{Fisheye}&
         FlashGS&$286$&$4.614$&$3034430$&$103.42$\\
           &   & \ &
           Vanilla&$213$&$17.101$&$13948372$&$1062.69$\\
         \cline{3-8} &   &
         \multirow{2}{*}{Pinhole}&
         FlashGS&$304$&$4.455$&$2415742$&$89.26$\\
           &   & \ &
           Vanilla&$189$&$15.000$&$13145841$&$1035.06$\\
         \hline
         \multirow{8}{*}{Kitchen} &
         \multirow{4}{*}{A100} &
         \multirow{2}{*}{Fisheye}&
         FlashGS&$570$&$4.038$&$2746724$&$93.36$\\
           &   & \ &
           Vanilla&$166$&$11.801$&$9834303$&$853.38$\\
         \cline{3-8} &   & 
         \multirow{2}{*}{Pinhole}&
         FlashGS&$534$&$3.793$&$2801239$&$94.61$\\
           &   & \ &
           Vanilla&$161$&$10.785$&$10945069$&$891.80$\\
         \cline{2-8} &
         % \multirow{4}{*}{V100} &
         % \multirow{2}{*}{Fisheye}&
         % Efficient&$446$&$4.276$&$2746724$&$93.36$\\
         %   &   & \ &
         %   Vanilla&$119$&$14.844$&$9834303$&$853.38$\\
         % \cline{3-8} &   & 
         % \multirow{2}{*}{Pinhole}&
         % Efficient&$549$&$3.629$&$2801239$&$94.61$\\
         %   &   & \ &
         %   Vanilla&$161$&$14.572$&$10945069$&$891.80$\\
         % \cline{2-8} &
         \multirow{4}{*}{RTX 4090} &
           \multirow{2}{*}{Fisheye}&
         FlashGS&$339$&$4.095$&$2746724$&$93.36$\\
           &   & \ &
           Vanilla&$282$&$12.002$&$9834303$&$853.38$\\
         \cline{3-8} &   &
         \multirow{2}{*}{Pinhole}&
         FlashGS&$342$&$4.137$&$2801239$&$94.61$\\
           &   & \ &
           Vanilla&$264$&$13.999$&$10945069$&$891.80$\\
    \end{tabular}
    \caption{Performance Comparison on different GPUs w/o using FlashGS. The "Intersects" in the table is the number of intersections between the Gaussians and the overlapped tiles. The result shows that
our method can adapt to Flash-GS in different GPUs, camera models, and scenes.}
    \label{performance}
\end{table}
\paragraph{Adaptation to Efficient Algorithms as Module} Compared by op43dgs\cite{huang2024erroranalysis3dgaussian}, Our method has only modified the geometry preprocessing and maintains other components as the vanilla 3DGS implements. This characteristic enables us to easily adapt Fisheye-GS as lightweight geometry preprocessing module to efficient rendering techniques like gsplat\cite{Ye_gsplat}, FlashGS\cite{feng2024flashgsefficient3dgaussian}, etc. We have applied our method to FlashGS, the efficient rendering pipeline for 3DGS which reduces the computation redundancies and optimizes the rasterizing process. As Table \ref{performance} shows, we have evaluated FPS, maximum rendering time per frame, the number of intersections between tiles and projected Gaussians, and memory consumption via different GPUs, rendering techniques, and camera models. The results show that
 the speedup in fisheye cameras is close to that in pinhole cameras on different settings. 

\begin{figure}[h]
    \centering
\includegraphics[width=\linewidth]{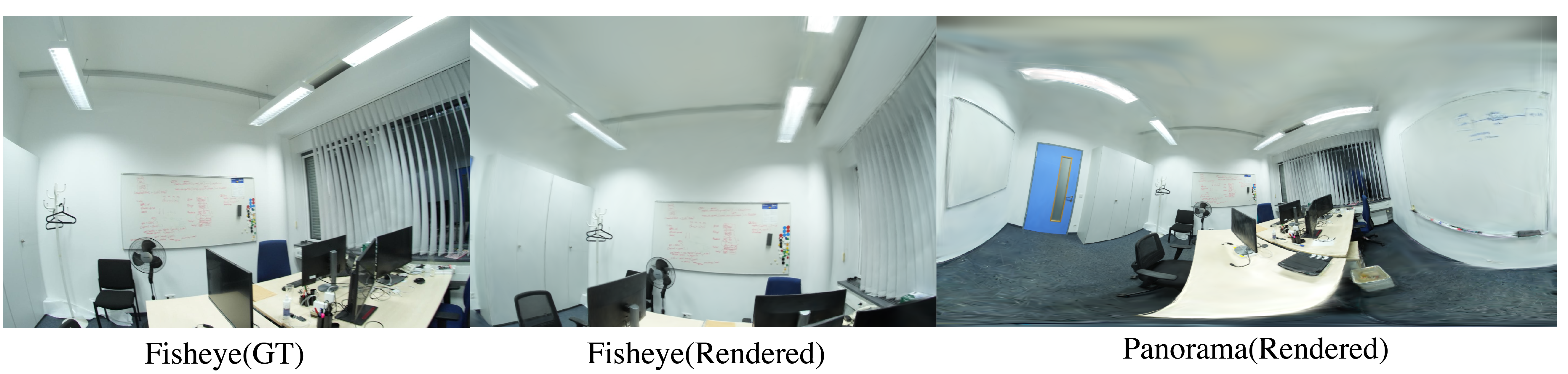}
    \caption{Comparison between rendered images in fisheye cameras and panorama cameras from the scene "Office Night" trained from Scannet++ dataset.}
    \vspace{-1em}
    \label{panorama}
\end{figure}

\paragraph{Extension to Other Cameras Models.} We have also extended our method to other projection transformation. For panorama cameras, we have modified the projection transformation $\phi$, the Jacobian matrix $\mathbf{J}$ and their gradients according to the ideal panorama cameras. Fig \ref{panorama} shows our method can both render the scenes with fisheye and panorama cameras. To extend our method to panorama cameras, we have to modify the projection as follows:
\begin{equation}
    x_p = \dfrac{w(\text{atan}_2(x_c,z_c) + \pi)}{2\pi}
\end{equation}
\begin{equation}
    y_p = \dfrac{h(\text{atan}_2(y_c,\sqrt{x_c^2+z_c^2)} + \frac{\pi}{2})}{\pi}
\end{equation}
\begin{equation}
    \mathbf{J}_\phi = \left[\begin{matrix}\frac{W z}{2 \pi \left(x^{2} + z^{2}\right)} & 0 & - \frac{W x}{2 \pi \left(x^{2} + z^{2}\right)}\\- \frac{H x y}{\pi \sqrt{x^{2} + z^{2}} \left(x^{2} + y^{2} + z^{2}\right)} & \frac{H \sqrt{x^{2} + z^{2}}}{\pi \left(x^{2} + y^{2} + z^{2}\right)} & - \frac{H y z}{\pi \sqrt{x^{2} + z^{2}} \left(x^{2} + y^{2} + z^{2}\right)}\end{matrix}\right]
\end{equation}
where $h, w$ is the height and width of the image, usually $w=2h$. 
\paragraph{Limitations.}Our method derives and implements the training and rendering process of 3DGS for fisheye camera models. We also evaluate the rendering quality of our method. However, our method ignores the approximation error of the projection for covariances. And because we explicitly compute the gradients, our method struggles to be generalized to other distortion models. Besides, our method is only based on ideal camera models but has not provided a projection model for generic cameras and a calibration method for real large-FOV cameras. Due to the lack of fisheye camera datasets, our method has not be tested by various types of scenes, like unbounded scenes, city-scale scenes, etc. 
\section{Conclusion} 
This paper introduces a new approach, Fisheye-GS, addressing the adaptability issues of 3D Gaussian Splatting (3DGS) across different camera models, particularly fisheye lenses. Fisheye-GS not only enables the reconstruction of 3DGS scenes captured by fisheye lenses but also functions as a lightweight module that can seamlessly integrate into high-performance 3DGS rendering algorithms. Additionally, it can be adapted for use with other camera models. This paper demonstrates that Fisheye-GS can effectively expand the application range of 3DGS technology to various camera perspectives, which is crucial for its broader real-life applications. We anticipate that the outcomes of this study will provide robust support and new perspectives for research and development in related fields.

\section*{Acknowledgment}
This work was supported by the National Key R$\&$D Program of China~(Grant No. 2022ZD0160201).

\bibliographystyle{splncs04}
\bibliography{main}

\begin{thebibliography}{10}
\providecommand{\url}[1]{\texttt{#1}}
\providecommand{\urlprefix}{URL }
\providecommand{\doi}[1]{https://doi.org/#1}

\bibitem{Bai2024360GSLP}
Bai, J., Huang, L., Guo, J., Gong, W., Li, Y., Guo, Y.: 360-{GS}: Layout-guided panoramic gaussian splatting for indoor roaming. arXiv preprint arXiv:2402.00763  (2024)

\bibitem{Barron2021MipNeRFAM}
Barron, J.T., Mildenhall, B., Tancik, M., Hedman, P., Martin-Brualla, R., Srinivasan, P.P.: Mip-{N}e{RF}: A multiscale representation for anti-aliasing neural radiance fields. In: Proceedings of the IEEE/CVF International Conference on Computer Vision. pp. 5855--5864 (2021)

\bibitem{Barron2021MipNeRF3U}
Barron, J.T., Mildenhall, B., Verbin, D., Srinivasan, P.P., Hedman, P.: Mip-{N}e{RF} 360: Unbounded anti-aliased neural radiance fields. In: Proceedings of the IEEE/CVF Conference on Computer Vision and Pattern Recognition. pp. 5470--5479 (2022)

\bibitem{Bourke2009idome}
Bourke, P.: idome: Immersive gaming with the {U}nity3{D} game engine. Computer Games and Allied Technology  \textbf{9} (2009)

\bibitem{Charatan2023pixelSplat3G}
Charatan, D., Li, S., Tagliasacchi, A., Sitzmann, V.: pixel{S}plat: 3{D} gaussian splats from image pairs for scalable generalizable 3{D} reconstruction. arXiv preprint arXiv:2312.12337  (2023)

\bibitem{Chen2024MVSplatE3}
Chen, Y., Xu, H., Zheng, C., Zhuang, B., Pollefeys, M., Geiger, A., Cham, T.J., Cai, J.: {MVS}plat: Efficient 3{D} gaussian splatting from sparse multi-view images. arXiv preprint arXiv:2403.14627  (2024)

\bibitem{deng2017cnn}
Deng, L., Yang, M., Qian, Y., Wang, C., Wang, B.: {CNN} based semantic segmentation for urban traffic scenes using fisheye camera. In: 2017 IEEE Intelligent Vehicles Symposium (IV). pp. 231--236. IEEE (2017)

\bibitem{feng2024flashgsefficient3dgaussian}
Feng, G., Chen, S., Fu, R., Liao, Z., Wang, Y., Liu, T., Pei, Z., Li, H., Zhang, X., Dai, B.: Flashgs: Efficient 3d gaussian splatting for large-scale and high-resolution rendering (2024), \url{https://arxiv.org/abs/2408.07967}

\bibitem{Yu2021PlenoxelsRF}
Fridovich-Keil, S., Yu, A., Tancik, M., Chen, Q., Recht, B., Kanazawa, A.: Plenoxels: Radiance fields without neural networks. In: Proceedings of the IEEE/CVF Conference on Computer Vision and Pattern Recognition. pp. 5501--5510 (2022)

\bibitem{huang2024erroranalysis3dgaussian}
Huang, L., Bai, J., Guo, J., Li, Y., Guo, Y.: On the error analysis of 3d gaussian splatting and an optimal projection strategy (2024), \url{https://arxiv.org/abs/2402.00752}

\bibitem{Jakab2024simulation}
Jakab, D., Deegan, B.M., Sharma, S., Grua, E.M., Horgan, J., Ward, E., Ven, P.v.d., Scanlan, A., Eising, C.: Surround-view fisheye optics in computer vision and simulation: Survey and challenges. IEEE Transactions on Intelligent Transportation Systems p. 1–22 (2024)

\bibitem{Jeong2021SelfCalibratingNR}
Jeong, Y., Ahn, S., Choy, C., Anandkumar, A., Cho, M., Park, J.: Self-calibrating neural radiance fields. In: Proceedings of the IEEE/CVF International Conference on Computer Vision. pp. 5846--5854 (2021)

\bibitem{3DGS}
Kerbl, B., Kopanas, G., Leimkuehler, T., Drettakis, G.: 3{D} gaussian splatting for real-time radiance field rendering. ACM Transactions on Graphics (TOG)  \textbf{42},  1--14 (2023)

\bibitem{konrad2024high}
Konrad, J., Cokbas, M., Ishwar, P., Little, T.D., Gevelber, M.: High-accuracy people counting in large spaces using overhead fisheye cameras. Energy and Buildings  \textbf{307},  113936 (2024)

\bibitem{Kulkarni2022360FusionNeRFPN}
Kulkarni, S., Yin, P., Scherer, S.: 360{F}usion{N}e{RF}: Panoramic neural radiance fields with joint guidance. In: 2023 IEEE/RSJ International Conference on Intelligent Robots and Systems (IROS). pp. 7202--7209. IEEE (2023)

\bibitem{Li2024DNGaussianOS}
Li, J., Zhang, J., Bai, X., Zheng, J., Ning, X., Zhou, J., Gu, L.: {DNG}aussian: Optimizing sparse-view 3{D} gaussian radiance fields with global-local depth normalization. arXiv preprint arXiv:2403.06912  (2024)

\bibitem{Liang2024AnalyticSplattingA3}
Liang, Z., Zhang, Q., Hu, W., Feng, Y., Zhu, L., Jia, K.: Analytic-splatting: Anti-aliased 3{D} gaussian splatting via analytic integration. arXiv preprint arXiv:2403.11056  (2024)

\bibitem{Lu2023ScaffoldGSS3}
Lu, T., Yu, M., Xu, L., Xiangli, Y., Wang, L., Lin, D., Dai, B.: Scaffold-{GS}: Structured 3{D} gaussians for view-adaptive rendering. arXiv preprint arXiv:2312.00109  (2023)

\bibitem{Mildenhall2020NeRFRS}
Mildenhall, B., Srinivasan, P.P., Tancik, M., Barron, J.T., Ramamoorthi, R., Ng, R.: Ne{RF}: Representing scenes as neural radiance fields for view synthesis. Communications of the ACM  \textbf{65}(1),  99--106 (2021)

\bibitem{Ren2024OctreeGSTC}
Ren, K., Jiang, L., Lu, T., Yu, M., Xu, L., Ni, Z., Dai, B.: Octree-{GS}: Towards consistent real-time rendering with lod-structured 3{D} gaussians. arXiv preprint arXiv:2403.17898  (2024)

\bibitem{schonberger2016structure}
Schonberger, J.L., Frahm, J.M.: Structure-from-motion revisited. In: Proceedings of the IEEE conference on computer vision and pattern recognition. pp. 4104--4113 (2016)

\bibitem{Shen2022OmniNeRFHO}
Shen, J., Song, B., Wu, Z., Xu, Y.: Omni{N}e{RF}: Hybriding omnidirectional distance and radiance fields for neural surface reconstruction. In: 2022 2nd International Conference on Computational Modeling, Simulation and Data Analysis (CMSDA). pp. 281--285. IEEE (2022)

\bibitem{TanciWNLYKWKASAMK2023}
Tancik, M., Weber, E., Ng, E., Li, R., Yi, B., Wang, T., Kristoffersen, A., Austin, J., Salahi, K., Ahuja, A., et~al.: Ne{RF}studio: A modular framework for neural radiance field development. In: ACM SIGGRAPH 2023 Conference Proceedings. pp. 1--12 (2023)

\bibitem{vandewiele2012visibility}
Vandewiele, F., Motamed, C., Yahiaoui, T.: Visibility management for object tracking in the context of a fisheye camera network. In: 2012 Sixth International Conference on Distributed Smart Cameras (ICDSC). pp.~1--6. IEEE (2012)

\bibitem{Wang2023PERFPN}
Wang, G., Wang, P., Chen, Z., Wang, W., Loy, C.C., Liu, Z.: {PERF}: Panoramic neural radiance field from a single panorama. arXiv preprint arXiv:2310.16831  (2023)

\bibitem{wang2004image}
Wang, Z., Bovik, A.C., Sheikh, H.R., Simoncelli, E.P.: Image quality assessment: from error visibility to structural similarity. IEEE transactions on image processing  \textbf{13}(4),  600--612 (2004)

\bibitem{Xian2023NeuralLM}
Xian, W., Bo{\v{z}}i{\v{c}}, A., Snavely, N., Lassner, C.: Neural lens modeling. In: Proceedings of the IEEE/CVF Conference on Computer Vision and Pattern Recognition. pp. 8435--8445 (2023)

\bibitem{Xiong2023SparseGSR3}
Xiong, H., Muttukuru, S., Upadhyay, R., Chari, P., Kadambi, A.: Sparse{GS}: Real-time 360° sparse view synthesis using gaussian splatting. arXiv e-prints pp. arXiv--2312 (2023)

\bibitem{xiong1997creating}
Xiong, Y., Turkowski, K.: Creating image-based {VR} using a self-calibrating fisheye lens. In: Proceedings of IEEE computer society conference on computer vision and pattern recognition. pp. 237--243. IEEE (1997)

\bibitem{Xu2022PointNeRFPN}
Xu, Q., Xu, Z., Philip, J., Bi, S., Shu, Z., Sunkavalli, K., Neumann, U.: Point-nerf: Point-based neural radiance fields. 2022 IEEE/CVF Conference on Computer Vision and Pattern Recognition (CVPR) pp. 5428--5438 (2022), \url{https://api.semanticscholar.org/CorpusID:246210101}

\bibitem{Yan2023MultiScale3G}
Yan, Z., Low, W.F., Chen, Y., Lee, G.H.: Multi-scale 3{D} gaussian splatting for anti-aliased rendering. arXiv preprint arXiv:2311.17089  (2023)

\bibitem{yang2023epformer}
Yang, Y., Deng, H.: {EP}former: an efficient transformer-based approach for retail product detection in fisheye images. Journal of Electronic Imaging  \textbf{32}(1),  013017--013017 (2023)

\bibitem{Ye_gsplat}
Ye, V., Kanazawa, A.: {gsplat}, \url{https://github.com/nerfstudio-project/gsplat}

\bibitem{ye2023mathematical}
Ye, V., Kanazawa, A.: Mathematical supplement for the $\texttt{gsplat}$ library. arXiv preprint arXiv:2312.02121  (2023)

\bibitem{yeshwanthliu2023scannetpp}
Yeshwanth, C., Liu, Y.C., Nie{\ss}ner, M., Dai, A.: Scan{N}et++: A high-fidelity dataset of 3{D} indoor scenes. In: Proceedings of the IEEE/CVF International Conference on Computer Vision. pp. 12--22 (2023)

\bibitem{Yu2021PlenOctreesFR}
Yu, A., Li, R., Tancik, M., Li, H., Ng, R., Kanazawa, A.: Plenoctrees for real-time rendering of neural radiance fields. In: Proceedings of the IEEE/CVF International Conference on Computer Vision. pp. 5752--5761 (2021)

\bibitem{Yu2023MipSplattingA3}
Yu, Z., Chen, A., Huang, B., Sattler, T., Geiger, A.: Mip-{S}platting: Alias-free 3{D} gaussian splatting. arXiv preprint arXiv:2311.16493  (2023)

\bibitem{zhang2018unreasonable}
Zhang, R., Isola, P., Efros, A.A., Shechtman, E., Wang, O.: The unreasonable effectiveness of deep features as a perceptual metric. In: Proceedings of the IEEE conference on computer vision and pattern recognition. pp. 586--595 (2018)

\bibitem{zhang2016meusem}
Zhang, S., Zhao, W., Wang, J., Luo, H., Feng, X., Peng, J.: A mixed-reality museum tourism framework based on {HMD} and fisheye camera. In: Proceedings of the 15th ACM SIGGRAPH Conference on Virtual-Reality Continuum and Its Applications in Industry-Volume 1. pp. 47--50 (2016)

\bibitem{Zhu2023FSGSRF}
Zhu, Z., Fan, Z., Jiang, Y., Wang, Z.: {FSGS}: Real-time few-shot view synthesis using gaussian splatting. arXiv preprint arXiv:2312.00451  (2023)

\end{thebibliography}

\end{document}